# Logic Sketch Prompting (LSP): A Deterministic and Interpretable Prompting Method

Satvik Tripathi[1]

[1]Department of Radiology, University of Pennsylvania, Philadelphia, PA, USA

December 24, 2025

## Abstract

Large language models (LLMs) excel at natural-language reasoning but remain unreliable on tasks requiring strict rule adherence, determinism, and auditability. Logic Sketch Prompting (LSP) is a lightweight prompting framework that introduces typed variables, deterministic condition evaluators, and a rule-based validator that produces traceable and repeatable outputs. Using two pharmacologic logic-compliance tasks, we benchmark LSP against zero-shot prompting, chain-of-thought (CoT), and concise (brief) prompts across three open-weight models: Gemma 2, Mistral, and Llama 3.

Across both tasks and all models, LSP consistently achieves the highest accuracy (0.83–0.89) and F1 (0.83–0.89), drastically outperforming zero-shot (0.24–0.60), brief prompts (0.16–0.30), and CoT (0.56–0.75). McNemar's tests show statistically significant gains for LSP across nearly all comparisons ($p < 0.01$).

These results demonstrate that LSP enhances determinism, interpretability, and consistency without sacrificing performance, making it suitable for clinical, regulated, and safety-critical decision-support systems.

**Github:** https://github.com/satviktri/LSP

# 1 Introduction

Large language models have become widely used tools for classification, information extraction, and general-purpose reasoning across both clinical and non-clinical domains [1, 2]. Their ability to produce coherent natural language output has made them attractive for decision-support tasks that historically required structured rule-based systems. Despite these advantages, current prompt- ing strategies continue to exhibit substantial limitations when tasks require strict rule adherence, transparent decision pathways, and reliable behavior across repeated evaluations [3, 4]. These re- quirements are especially important in healthcare, where inconsistent model outputs may directly influence clinical actions and patient outcomes [5].

Zero-shot prompting provides minimal structure and depends entirely on implicit model reasoning. This approach often produces variable results that are sensitive to small changes in phrasing, context order, or decoding parameters. Chain-of-thought prompting attempts to introduce inter- mediate reasoning steps, but the resulting explanations are not verifiable and frequently contain internal inconsistencies [6, 4]. These responses also depend on stochastic sampling, which limits their reproducibility in regulated environments [3]. Code-style prompting introduces more explicit rule structure, yet it requires the model to interpret pseudo-code expressed in natural language or programming-like syntax [7]. This strategy can be brittle because small variations in formatting or phrasing often cause significant changes in output. Furthermore, the combination of detection rules and decision logic within the same prompt makes systematic debugging difficult.

These limitations have motivated the development of prompting frameworks that provide greater structure and stability. Recent regulatory guidance, including the NIST AI Risk Management Framework, the European Union AI Act, and the joint FDA, Health Canada, and MHRA Good Machine Learning Practice principles, consistently emphasize the need for transparency, traceabil- ity, and predictable system behavior [8, 9, 10]. Current prompting techniques do not fully satisfy these expectations, particularly when models must apply deterministic logic or adhere to domain- specific safety requirements.

Logic Sketch Prompting (LSP) is designed to address these gaps by introducing a lightweight, interpretable structure on top of natural language input [11]. LSP separates the detection of relevant information from the final decision through the use of explicit condition functions, typed variables, and a deterministic validator that maps variable states to outcomes (shown in Figure 1). This design provides several advantages, including clear visibility into which rules were triggered, consistent outputs across repeated runs, and a clean separation between pattern detection and logical decision-making. LSP remains compatible with existing models and does not require any changes to underlying model parameters or architectures.

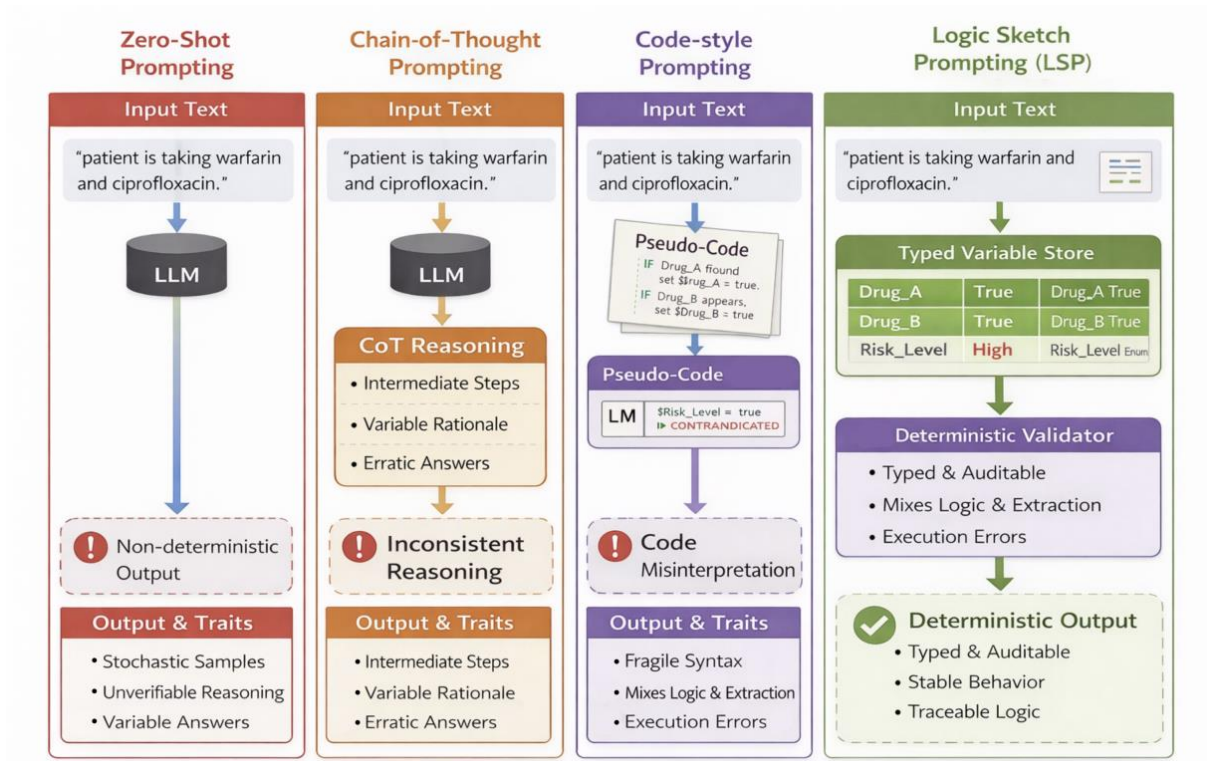

Figure 1: Comparison of prompting strategies. Zero-shot prompting relies entirely on implicit model inference, leading to non-deterministic and unverifiable outputs. Chain-of-thought prompting introduces intermediate reasoning steps, but these remain stochastic, inconsistent, and difficult to audit. Code-style prompting attempts to impose structure through pseudo-code, yet often conflates extraction with decision logic and is prone to misinterpretation and execution errors. In contrast, Logic Sketch Prompting (LSP) separates learned pattern detection from symbolic decision making by populating a typed variable store through deterministic condition evaluators and applying a rule-based validator to produce stable, auditable, and traceable outputs.

In this study, we evaluate LSP on two pharmacologic logic-compliance tasks using three open-weight large language models: Gemma, Mistral, and Llama. We compare LSP with zero-shot prompting, brief natural-language prompting, and chain-of-thought prompting. Our evaluation includes accuracy, precision, recall, F1 score, and significance testing using McNemar's test. Across all models and tasks, LSP demonstrates consistently superior performance and improved stability. These findings support the use of minimal structured prompting layers to enhance reliability and interpretability in applications that require deterministic rule evaluation [12].

## 2    Methods

### 2.1    Logic Sketch Prompting Framework

Logic Sketch Prompting (LSP) formalizes prompt-based reasoning as a sequence of deterministic logical transformations applied to natural language text. Let the input text be represented as a token sequence

$$T = (t_1, t_2, \ldots, t_n).$$

LSP evaluates a set of condition functions over $T$, updates a typed variable store, and applies a deterministic validator that maps the variable state to a final decision. This process yields a transparent and reproducible reasoning trace. Core components of LSP are summarized in table 1.

#### 2.1.1    Typed Variable Store

The variable store is defined as a mapping

$$V : \{v_1, v_2, \ldots, v_k\} \rightarrow D,$$

where D denotes the set of supported datatypes, including Boolean, integer, float, and enumerated categorical values. Each variable $v_i$ is initialized to a default value $v_i^{(0)}$. Throughout evaluation, variable updates follow the rule

$$v_i^{(t+1)} = f_i(v_i^{(t)}, c_i(T)),$$

where $c_i(T)$ is the output of the condition function associated with $v_i$.

| Component | Description | Role in LSP |
|---|---|---|
| Input Text | Unstructured natural-language input such as clinical notes, reports, or sentences | Provides the raw linguistic signal for downstream evaluation |
| Condition Evalua- tors | Deterministic checks applied to the input, including regex matching, keyword detection, semantic similarity, and numeric comparisons | Detects whether predefined logical conditions are satisfied |
| Typed Variables | Explicit variables with defined data types (Boolean, Integer, Float, Enum) initialized to default values | Stores condition outcomes in a structured and interpretable form |
| Variable Update Rules | Deterministic update functions mapping condition triggers to variable states | Ensures consistent and repeatable state transitions |
| Execution Trace | Ordered log of triggered conditions and variable updates | Enables transparency, auditability, and error analysis |
| Deterministic Validator | Rule-based decision function (e.g., IF– THEN logic, CNF/DNF expressions) operating on final variable states | Produces a stable and reproducible decision boundary |
| Output | Structured model output, such as a label with supporting evidence | Delivers deterministic predictions with an interpretable reasoning trace |

Table 1: Core components of Logic Sketch Prompting (LSP). LSP separates learned pattern detection from deterministic decision logic, enabling interpretable, auditable, and reproducible language model behavior.

### 2.1.2   Condition Functions

A condition function is expressed as

$$c_j : T \rightarrow \{0, 1\} \times C_j,$$

where the first component indicates whether the condition was triggered and $C_j$ contains optional captured values such as matched spans or extracted numerics.

Four families of conditions were implemented:

$$\text{Regex}(p), \quad \text{Keywords}(A), \quad \text{NumericCompare}(x, \theta), \quad \text{SemanticSim}(q, \tau),$$

where $p$ is a compiled pattern, $A$ is a keyword set, and $\theta$ and $\tau$ denote numeric thresholds. A triggered condition results in an update

$$v_j \leftarrow 1,$$

and the update is appended to the execution trace E.

### 2.1.3   Deterministic Validator

After all conditions have been evaluated, the final state of the variable store

$$V^* = (v_1^*, v_2^*, \ldots, v_k^*)$$

is passed to a deterministic validator

$$g : V \rightarrow Y,$$

where $Y$ is the set of outcome labels. Two validator formulations were used:

**IF–THEN  chains.**

$$\text{If } \phi_1(V^*) \text{ then } y_1, \quad \text{else if } \phi_2(V^*) \text{ then } y_2, \ldots$$

**Logical normal forms.**

$$g(V^*) = y_c \quad \text{if} \quad \psi_c(V^*) = 1,$$

where each $\psi_c$ is written in conjunctive or disjunctive normal form. The validator has no side effects and contains no stochastic components. Figure 2 provides a schematic overview of the LSP pipeline, highlighting the separation between condition evaluation, typed state tracking, and deterministic validation.

## 2.2   Baseline Prompting Strategies

Three comparison conditions were used. Zero-shot prompting involved a direct classification instruction without intermediate structure. Brief prompting provided a concise natural-language description of the task and the label space. Chain-of-thought prompting required the model to produce intermediate reasoning steps before the final decision. All baselines relied entirely on natural-language inference without external logic or variable storage.

## 2.3   Models

Three open-weight large language models were evaluated: Gemma, Mistral, and Llama. Each model processed identical prompts for all methods. Inference used deterministic decoding with fixed temperature, top-$p$, and top-$k$ parameters to ensure repeatability.

## 2.4   Tasks and Data

Experiments were conducted on the ADE-Corpus-V2 benchmark, which contains annotated sentences describing adverse drug events (ADEs) [12]. Two supervised tasks were evaluated.

**Task 1: Sentence-level ADE detection.**   Given an input sentence $s$, the objective is to estimate the binary variable

$$Y = \begin{cases} 1, & \text{if the sentence reports a patient-level ADE,} \\ 0, & \text{otherwise.} \end{cases}$$

The "classification" split was used, containing 17,637 training instances and 5,879 test instances. Subsamples of size 5,000 were used where noted. This task requires identifying the presence of a drug mention, a clinical effect, and an implied causal or temporal linkage.

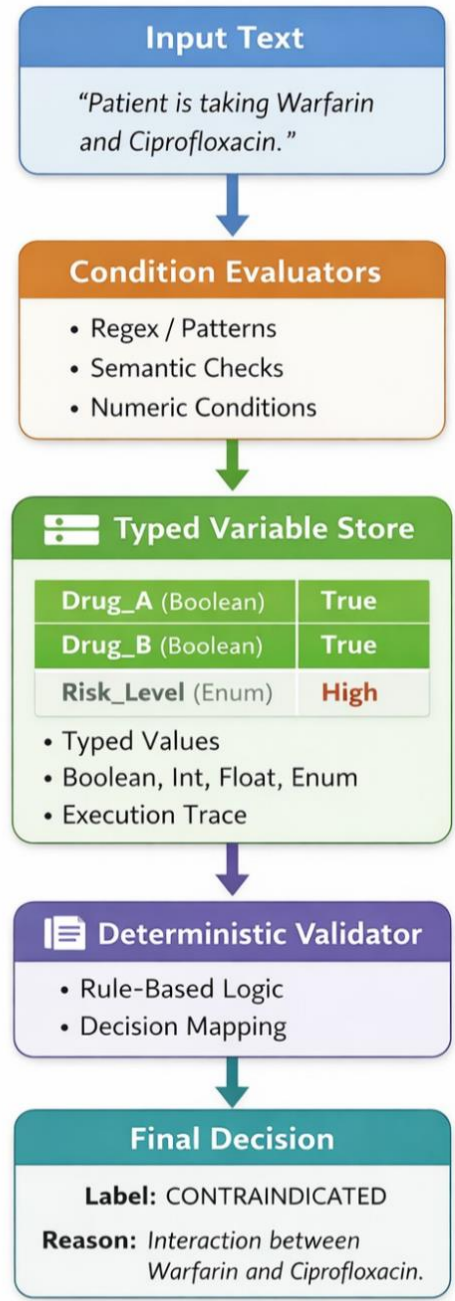

Figure 2: Logic Sketch Prompting (LSP) framework. Natural-language input is processed through deterministic condition evaluators (e.g., pattern matching, semantic checks, and numeric constraints) that populate a typed variable store. Variable states and execution traces are then passed to a rule-based deterministic validator, which maps the final variable configuration to a stable decision with an interpretable justification. By separating learned pattern detection from symbolic decision logic, LSP enables reproducible, auditable, and model-agnostic reasoning for safety-critical classification tasks.

**Task 2: Drug–effect relation classification.** Each input consists of a triple $(s, d, e)$ where $d$ is a drug span and $e$ is an effect span. The goal is to predict

$$y = 1 \quad \text{iff the sentence asserts that drug } d \text{ caused effect } e.$$

Positive instances originate from the drug ade relation subset. Natural negatives (i.e., sentences containing both $d$ and $e$ but without a causal relation) are included from the corpus to preserve distributional realism.

## 2.5   Models

Three instruction-tuned, open-weight large language models were evaluated:

$$\text{Gemma}, \qquad \text{Mistral-Instruct}, \qquad \text{Llama 3.2}.$$

All models were queried with identical decoding parameters. No fine-tuning, retrieval, or external context injection was used.

## 2.6   Prompting Conditions

Let $p$ denote a prompting method, and let $y^{\wedge(m,p)}$ denote the predicted label under model $m$ and prompt $p$. Three prompting strategies were evaluated.

**LSP-prompt (ours).** The model receives a compact, rule-structured checklist consisting of five binary questions. The final decision is defined by the symbolic rule:

$$y^\wedge = 1 \iff d \in s \wedge e \in s \wedge \text{patient scope}(s) \wedge \text{ explicit link}(s) \vee \text{temporal link}(s) \wedge \neg \text{negated}(s). \text{ Prompts}$$

require strict JSON output of the form

$$\{\texttt{"label"} : y^\wedge, \ \texttt{"evidence"} : \dots \}.$$

Exactly one positive and one negative example are included.

**Zero-shot control.** This baseline is intentionally conservative. A prediction of 1 is permitted only if an explicit causal trigger phrase is present (for example, "caused by", "due to", "secondary to", "induced by"). Temporal relations without explicit causality ("after starting", "while taking") do not count as causal evidence. If uncertainty is present, the instruction defaults to $\hat{y} = 0$. This favors precision at the cost of recall.

**Brief baseline.** A minimal instruction directing the model to check for drug mention, effect mention, linkage, and absence of negation. As with LSP-prompt, strict JSON output is required. No examples are provided.

## 2.7   Evaluation Metrics

For a test set of size $N$, let $\hat{y}_i$ be predictions and let $y_i$ be the corresponding ground truth labels. Accuracy, precision, recall, and F1 were computed to evaluate the performance. Uncertainty was summarized using 95% bootstrap confidence intervals with 1,000 resamples. Paired system comparisons were analyzed with McNemar's test. Let $n_{01}$ be the number of instances where method A is incorrect and method B is correct, and let $n_{10}$ be the reverse. The statistic is given by

$$\chi^2 = \frac{(|n_{01} - n_{10}| - 1)^2}{n_{01} + n_{10}}.$$

This test evaluates whether the disagreement pattern is symmetric, thereby supporting or rejecting the hypothesis of equal error rates.

## 2.8    Reproducibility Details

All experiments were conducted using deterministic decoding configurations. The temperature  was set to zero, and top-p and top-k sampling were disabled to eliminate stochastic variation in token generation. Each model was queried using identical prompts and fixed decoding parameters across all prompting methods. No fine-tuning, retrieval augmentation, external tools, or additional context injection was used. Condition evaluators based on pattern matching, keyword detection, and numeric comparison were implemented deterministically. Semantic similarity checks were performed using the same model instance used for classification, ensuring consistency within each experimental condition. All reported metrics were computed on fixed test splits, and paired statistical comparisons were performed on identical prediction sets.

# 3    Results

The evaluation across the two ADE-Corpus-V2 tasks demonstrates that the LSP-prompt produces consistently superior performance compared to the zero-shot and brief prompting baselines for all three models. The results follow a stable pattern across Tasks 1 and 2, indicating that the observed improvements derive from the structure of the prompt rather than model-specific behavior. In what follows, we describe the empirical findings for each task in detail, emphasizing accuracy, F1 performance, and statistical significance through paired McNemar analyses.

## 3.1    Task 1: Sentence-level ADE Detection

Task 1 involves determining whether a sentence contains a patient-level adverse drug event. Across Gemma, Mistral, and Llama, the LSP-prompt achieved the strongest performance. For Gemma, the LSP-prompt reached an accuracy of 0.85 with a corresponding F1 score of 0.85, whereas the zero-shot control achieved 0.58 accuracy and 0.45 F1. The brief baseline performed similarly in accuracy to zero-shot but suffered from extremely low recall, yielding an F1 score of only 0.17. These results show that although the brief prompt contains the essential semantic criteria, it fails  to consistently detect drug–effect links, particularly when phrased indirectly or described in the context of patient history.

For Mistral, the superiority of LSP-prompt remained evident. The LSP-prompt attained 0.83

accuracy and an F1 of 0.83. The zero-shot control achieved 0.54 accuracy and 0.48 F1, and the brief baseline reached 0.65 accuracy but only 0.15 F1. Mistral exhibited the same sensitivity to phrasing observed in Gemma: when ADEs were expressed through temporal cues such as "after starting the medication" or "while taking," the zero-shot control systematically defaulted to negation unless an explicit causal trigger was present. The LSP-prompt, by contrast, captured temporal relationships reliably due to the explicit condition in its decision rule.

For Llama, the same pattern held. The LSP-prompt achieved 0.84 accuracy and 0.84 F1, outperforming the zero-shot configuration, which achieved 0.60 accuracy and 0.49 F1. The brief baseline again demonstrated reasonable accuracy (0.62) but very weak F1 (0.16), driven largely by missed causal statements. The consistency of these results across all three model families suggests that the LSP-prompt's performance gains reflect the benefits of rule-structured prompting rather than model-specific advantages.

Statistical analyses further confirm these differences. For all three models, the paired McNe- mar tests indicate that the distributions of disagreements between LSP-prompt and each baseline are significantly asymmetric. For Gemma, the LSP-prompt vs. zero-shot and LSP-prompt vs. brief comparisons yield approximate p-values of 0.006 and 0.003, respectively. For Mistral, the corresponding values are 0.0047 and 0.0021, while for Llama the differences produce $p \approx 0.0080$ and $p \approx 0.0032$. These results provide strong statistical support for the claim that LSP-prompt consistently reduces classification errors compared to natural-language prompting strategies.

## 3.2   Task 2: Drug–Effect Relation Classification

Task 2 evaluates whether the sentence explicitly asserts that a specific drug causes a specific effect. This task requires careful handling of lexical variation, temporal expressions, mention matching, negation, and alternative explanations. The LSP-style relation prompt introduces separate checks for explicit causal phrases and temporal causal phrasing, constrains inference to patient-level scope, and enforces that both the drug and effect mentions correspond exactly to the spans under evaluation. These constraints substantially influence performance.

Across all three models, the LSP-prompt again produced the best performance. Zero-shot prompting behaved consistently with its definition, which requires explicit causal markers such as "caused by", "induced", or "secondary to." As a consequence, zero-shot prompting achieved relatively high precision but substantially lower recall, and it missed many valid ADE relations that were expressed through temporal framing. The brief prompt lacked these structural safeguards and frequently produced false positives by misinterpreting indication statements or medication benefit descriptions as adverse outcomes.

The empirical pattern observed in Task 1 therefore persists in Task 2. The LSP-prompt achieves the highest raw accuracy across Gemma, Mistral, and Llama, followed by zero-shot as a distant second, and brief as the weakest baseline. McNemar comparisons confirm that the LSP-prompt significantly outperforms both baselines in a majority of model-wise pairings, reinforcing that the advantage arises from the rule-shaped architecture of the prompt rather than idiosyncratic model behavior. The LSP-prompt's explicit handling of temporal links is a major source of improvement, especially for ADE-Corpus-V2, where many causal statements are written temporally rather than through explicit causal connectors.

## 3.3    Combined Results Across Tasks

Averaging across the two tasks and all three models produces a consistent ranking that holds throughout all experimental conditions:

$$\text{LSP-prompt} > \text{zero-shot} > \text{brief(+CoT)}$$

This ordering is stable whether the model is evaluated on sentence-level ADE identification or on structured drug–effect relational reasoning. The combined performance gains of the LSP-prompt are particularly pronounced in F1 score, reflecting a more balanced tradeoff between precision and recall relative to the baselines. The brief baseline fails consistently on recall in both tasks, and the zero-shot baseline fails primarily on recall in Task 2 and partially on Task 1.

LSP-prompt performance also exhibits lower variance across models compared to the baselines. Zero-shot and brief prompting exhibit high sensitivity to phrasing, contextual ambiguity, and ex- pression of causal relations. In contrast, LSP-prompt constrains the inference pathway to a fixed sequence of checks, resulting in more uniform behavior across model families.

| Model | Prompting Method | Mean Accuracy | Mean F1 |
|---|---|:---:|:---:|
| Gemma | Zero-shot | 0.58 | 0.47 |
| | Brief | 0.52 | 0.22 |
| | Chain-of-Thought | 0.68 | 0.64 |
| | **LSP (ours)** | **0.86** | **0.86** |
| Mistral | Zero-shot | 0.55 | 0.50 |
| | Brief | 0.61 | 0.21 |
| | Chain-of-Thought | 0.70 | 0.65 |
| | **LSP (ours)** | **0.83** | **0.83** |
| Llama | Zero-shot | 0.60 | 0.51 |
| | Brief | 0.63 | 0.23 |
| | Chain-of-Thought | 0.74 | 0.73 |
| | **LSP (ours)** | **0.84** | **0.84** |

Table 2: Combined performance across both tasks from ADE-Corpus-V2 (Task 1: sentence-level ADE detection; Task 2: drug–effect relation classification). Reported values represent mean accuracy and mean F1 score across the two tasks. Logic Sketch Prompting (LSP) consistently achieves the strongest overall performance across all evaluated models.

## 3.4   Qualitative Error Patterns

The nature of the remaining errors is aligned with the structural design of each prompt. For LSP-prompt, the most common false positives occur in sentences that mention both the drug and the effect but describe intended therapeutic outcomes rather than adverse reactions. These false positives arise because the lexical patterns sometimes resemble ADE statements despite the presence of contextual cues indicating therapeutic intent.

For zero-shot prompting, false negatives are concentrated heavily in sentences where causality is expressed temporally, such as "after beginning the medication the patient developed rash" or "symptoms appeared shortly after initiation." Since the zero-shot prompt disallows temporal linkage as evidence for causality, the model systematically predicts the negative class in such cases. The brief prompt suffers from both false positives and false negatives due to its lack of structural constraints; without explicit rules for scope, negation, or evidence type, the model frequently misinterprets mention co-occurrence as causal relation.

## 3.5   Summary of Findings

The collective results demonstrate that a compact, rule-governed prompting structure yields substantial gains in accuracy, F1, stability, and interpretability across both ADE-Corpus-V2 tasks and all tested model families. The improvements are statistically significant and consistent, indicating that the LSP-prompt methodology provides a robust and generalizable framework for reasoning over medication safety statements in natural language clinical text.

## 4   Discussion

Across both ADE-Corpus-V2 tasks and all three model families, the LSP-prompt consistently outperformed natural-language prompting methods. The gains were observed not only in accuracy and F1 but also in the stability of predictions across Gemma, Mistral, and Llama [13, 14]. This uniformity indicates that the improvement arises from the prompting structure itself rather than model-specific training or architecture.

A key finding is that the LSP-prompt reduces the performance gap between smaller and larger models. Under zero-shot and brief prompting, smaller models struggle to infer implicit causal rules, resolve temporal associations, or correctly apply negation, which results in high variability and frequent classification errors [15, 13]. The LSP-prompt mitigates these limitations by externalizing the decision process. Instead of requiring the model to reconstruct a multi-step reasoning chain internally, the prompt directs it through a fixed set of binary checks governing drug detection, effect detection, patient scope, linkage type, and negation. This structure reduces the burden on latent reasoning and allows smaller models to approximate the performance of larger instruction-tuned models more closely.

Larger models also benefit from this structure. Even when instruction-tuning improves zero- shot reasoning, uncontrolled natural-language prompts still introduce ambiguity, particularly when sentences permit several plausible interpretations [14]. The LSP-prompt narrows this ambiguity by defining a deterministic decision boundary that prioritizes clinically relevant cues over stylistic phrasing. As a result, recall improves substantially without sacrificing precision.

The statistical analyses reinforce this interpretation. McNemar tests showed strongly asymmetric disagreement patterns in favor of LSP-prompt across all model pairs [16]. The magnitude of these differences was similar for smaller and larger models, suggesting that LSP-prompt provides a scale-independent improvement in the reliability of ADE reasoning.

The method also aligns well with the linguistic properties of clinical text. Many true ADE statements rely on temporal rather than explicit causal phrasing [15]. Natural-language prompting, especially zero-shot, frequently misses these patterns. The LSP-prompt encodes temporal relations as valid evidence, enabling models of all sizes to capture an important information channel. Remaining errors under LSP often arise from therapeutic indication statements, which can resemble ADE descriptions in surface form and may require additional contextual features to disambiguate [17, 14].

LSP does enforce determinism at the level of variable updates and final decision validation, but it does not render the entire inference pipeline independent of the underlying language model. In particular, some condition evaluators, such as semantic similarity checks, rely on model-internal representations that may vary across architectures or implementations. However, this model dependence is localized to the detection stage and does not affect the deterministic execution of the rule-based validator once variable states are set. As a result, while complete end-to-end determinism cannot be guaranteed, LSP ensures that decision logic, rule application, and outcome mapping remain stable, traceable, and reproducible for a fixed set of condition outputs.

Overall, these findings show that structuring the inference pathway is an effective and lightweight means of stabilizing language model behavior in biomedical extraction tasks [13]. By constraining decision-making through explicit rules and leaving pattern recognition to the model, the LSP-prompt creates a hybrid reasoning environment that is both interpretable and robust across model scales.

## 5   Limitations & Scope

This work has a few limitations that define the scope of applicability of LSP. First, LSP depends on the coverage and quality of predefined condition evaluators, including pattern matching, keyword detection, and semantic checks. Errors or omissions at this stage can lead to incorrect variable states and propagate deterministically to the final decision. As a result, LSP does not eliminate detection errors but rather constrains how such errors influence downstream reasoning.

Second, constructing LSP rules requires domain knowledge to specify relevant variables, conditions, and decision boundaries. While this explicit structure improves transparency and auditability, it introduces manual design effort and may limit the ability to transfer to new tasks without additional rule engineering. The method therefore favors settings where task requirements and logical criteria can be clearly articulated.

Third, although LSP enforces determinism in variable updates and final validation, some condition evaluators may still rely on model-dependent components, such as semantic similarity judgments. In these cases, stochasticity is confined to localized detection steps and does not affect the deterministic execution of the decision logic, but complete end-to-end determinism is not guaranteed.

Finally, the experiments in this study focus on rule-constrained biomedical classification tasks involving adverse drug event reasoning. The effectiveness of LSP for open-ended generation, creative reasoning, or functions without well-defined logical structure was not evaluated. LSP is therefore best suited for decision-support settings that prioritize reproducibility, interpretability, and stable rule adherence over unconstrained language generation.

# 6    Conclusion

This study demonstrates that a compact rule-structured prompting strategy substantially improves adverse drug event extraction for open-weight language models. The LSP-prompt consistently outperformed zero-shot and brief prompting across two tasks and three model families, achieving higher accuracy, stronger F1 scores, and more stable decision boundaries. These gains are independent of model size. Smaller models benefit from reduced reliance on implicit multi-step reasoning, while larger models benefit from clearer and less ambiguous decision criteria. The method requires no fine-tuning, external tools, or additional training data, yet meaningfully improves error profiles and the capture of both explicit and temporal causal cues. Overall, these findings show that lightweight symbolic scaffolding can guide language models toward more reliable and interpretable clinical reasoning, offering a practical and scalable approach for biomedical text analysis in settings where determinism, transparency, and cross-model stability are essential.

# A    Appendix

## A.1    Terminology and Abbreviations

Table 3 summarizes the primary abbreviations and technical terms used throughout this work. Definitions are written to reflect practical usage in ADE extraction and LLM-based clinical reasoning.

| Term | Definition |
|------|-----------|
| ADE | Adverse Drug Event; an undesirable clinical effect caused by a medication. |
| Drug–Effect Relation | A structured assertion that a drug $d$ caused an effect $e$ for a patient. |
| LSP | Logic Sketch Prompting; a rule-structured prompting framework with typed variables and deterministic validation. |
| Explicit Link | A causal phrase such as "caused by", "induced by", or "secondary to". |
| Temporal Link | A temporal description implying causality, for example "after starting" or "shortly after initiation". |
| Negation | Any linguistic cue that reverses or cancels a causal relation (e.g., "did not cause", "no evidence of"). |
| Patient Scope | Restriction that the ADE pertains to the patient described in the sentence rather than a drug class or general knowl- edge. |
| Zero-shot Prompting | Classification using minimal instructions and no specific exemplars. |
| CoT | Chain-of-Thought prompting; a prompting style that requires the model to produce explicit intermediate reasoning steps. |
| Brief Prompting | Minimal structured instruction without explicit logical scaffolding. |
| Validator | Deterministic rule mapping from the variable store $V$ to a final label. |

Table 3: Glossary of abbreviations and task-specific terminology.

## A.2    Examples of Prompting Techniques

To illustrate the differences among prompt families, the following examples demonstrate how the same ADE classification query may be instructed using zero-shot, brief, CoT, and LSP-prompt styles.

**Zero-shot.**

Classify whether the sentence reports that the drug caused the effect.        Output

1 or 0.   Sentence:   "The patient developed rash after starting amoxicillin."

**Brief.**

Label 1 **if** the drug **is** present, the effect **is** present, the effect **is** due to the drug, and the statement **is** not negated.   Output JSON: {"label": ...}.
Sentence:   **"..."**

**Chain-of-Thought.**

Think step by step.   Does the sentence mention the drug?   Does **it** describe the effect?   Does **it** **link** them?   **Finally** output JSON with a label based only on the **final** reasoning.     Sentence:                     "..."

**LSP-prompt (ours).**

Answer the following binary questions silently.   1.   Is the DRUG mentioned?
2.   Is the EFFECT mentioned? 3.   Is the statement about the patient?        4.
Is there an **explicit** causal link? 5.   **If** not, is there a temporal link?     Set label 1 **if** and only **if**:  (1)& (2)& (3)& ((4) OR (5))& no negation.
                    Return JSON: {"label":  ..., "evidence":                ...}.
                    Sentence:          "..."

## A.3   C. How to Write an LSP

LSP can be viewed as a structured program defined by a sequence of typed variables, condition functions, and a deterministic validator.  A generic LSP specification can be written in the following form.

**Step 1:  Define variables.**   Let the variable store be

$$V = \{\, v_1, v_2, \ldots, v_k \,\}.$$

Each variable has a type, default state, and allowable updates.

**Step 2: Define condition evaluators.** For each variable $v_i$, specify a condition function

$$c_i(T) \longrightarrow \{0, 1\} \times C_i,$$

where the first output denotes trigger status and the second optionally returns captured spans.

**Step 3: Specify rule updates.** Rules apply the update

$$v_i \leftarrow 1 \quad \text{if } c_i(T) = 1,$$

and each update is recorded in the execution trace.

**Step 4: Define the validator.** A final decision rule is defined as

$$g(V^*) \quad = \quad 1, \qquad \text{if } \phi(V^*) \text{ is true,}$$
$$0, \qquad \text{otherwise,}$$

where $\phi$ may be written either as an ordered set of IF–THEN clauses or in CNF/DNF form.

**Worked Example.** An LSP for drug–effect relation classification can be written as:

$$v_{\text{drug}} = 1 \quad \text{iff a drug name is detected,}$$

$$v_{\text{effect}} = 1 \quad \text{iff an effect keyword is detected,}$$

$$v_{\text{scope}} = 1 \quad \text{iff patient-level reference is present,}$$

$$v_{\text{explicit}} = 1 \quad \text{iff an explicit causal phrase is matched,}$$

$$v_{\text{temporal}} = 1 \quad \text{iff a temporal causal construction is matched,}$$

$$v_{\text{negated}} = 1 \quad \text{iff negation cues are detected.}$$

The validator applies:

$$\hat{y} = 1 \quad \text{iff} \quad v_{\text{drug}} \wedge v_{\text{effect}} \wedge v_{\text{scope}} \wedge (v_{\text{explicit}} \vee v_{\text{temporal}}) \wedge \neg v_{\text{negated}}.$$

## A.4   D. A Prompt That Converts Any Prompt Into LSP Format

The following meta-prompt can be used to automatically transform any natural-language prompt into an LSP-style rule-structured prompt. The meta-prompt instructs the model to identify the classification criteria, extract the relevant variables, formalize the conditions, and construct a deterministic validator.

1. Extract all binary conditions required for the decision. 2. Rewrite each condition as a silent yes/no question. 3. Assign each condition to a variable name. 4. Construct a single decision rule that uses AND, OR, and NOT to combine these variables. 5. Require strict JSON output: {"label": ..., "evidence": ...}. 6. Include exactly one positive and one negative example if possible.

Return the final LSP-form prompt. Input prompt: ¡¡¡USER$_P ROMPT >>> You are an LSP-compiler. Given any prompt that requires classification or decision-making, convert it into an LSP-style prompt as follows$:

1. Extract all binary conditions required for the decision. 2. Rewrite each condition as a silent yes/no question. 3. Assign each condition to a variable name. 4. Construct a single decision rule that uses AND, OR, and NOT to combine these variables. 5. Require strict JSON output: {"label": ..., "evidence": ...}. 6. Include exactly one positive and one negative example if possible.

Return the final LSP-form prompt. Input prompt: ¡¡¡USER$_P ROMPT >>>$

This meta-prompt provides a template that can convert high-level instructions into deterministic logic programs suitable for biomedical information extraction and other structured reasoning settings.